\documentclass[runningheads]{llncs}

\usepackage[T1]{fontenc}
\usepackage{graphicx}

\usepackage[colorlinks=true, linkcolor=blue, citecolor=blue, urlcolor=blue]{hyperref}

\begin{document}

\title{Role-Playing Evaluation for Large Language Models}

\author{Yassine El Boudouri \orcidID{0009-0001-4293-961X} \and
Walter Nuninger \orcidID{0000-0002-2639-1359} \and
Julian Alvarez \orcidID{0000-0002-9862-9485} \and
Yvan Peter \orcidID{0000-0002-1145-357X}}

\authorrunning{Yassine El Boudouri et al.}

\institute{Univ. Lille, CNRS, Centrale Lille, UMR 9189 CRIStAL, F-59000 Lille, France \email{\{yassine.el-boudouri,walter.nuninger,julian.alvarez,yvan.peter\}@univ-lille.fr}}

\sloppy
\maketitle 

\begin{abstract}
Large Language Models (LLMs) demonstrate a notable capacity for adopting personas and engaging in role-playing. However, evaluating this ability presents significant challenges, as human assessments are resource-intensive and automated evaluations can be biased. To address this, we introduce \textbf{Role-Playing Eval (RPEval)}, a novel benchmark designed to assess LLM role-playing capabilities across four key dimensions: emotional understanding, decision-making, moral alignment, and in-character consistency. This article details the construction of RPEval and presents baseline evaluations. Our code and dataset are available at \url{https://github.com/yelboudouri/RPEval}

\keywords{Role-playing  \and Large language models \and Benchmark \and Evaluation.}
\end{abstract}

\section{Introduction}
This study is part of the RVRC4.0 project, which develops digital educational resources for teaching soft skills essential to customer relationship management in industries such as retail, travel, and banking. The project focuses on core interpersonal competencies—including communication, decision-making, initiative, negotiation, and service orientation—that are vital in client-facing roles but often neglected in traditional training environments. Within the project, role-playing is adopted as a key pedagogical approach to support the development of interpersonal competencies. Learners participate in structured simulations of customer interactions—such as processing product returns, addressing complaints, or providing guidance—each designed to correspond with specific learning objectives. These scenarios aim to approximate real-world situations encountered in service-oriented professions, offering a context in which learners can apply and reflect on their soft skills.

While there is no universally accepted definition of role-playing, its interpretation often varies depending on the domain in which it is applied. In an academic context, role-playing is often described in ways that highlight its experiential aspects. For instance, Sellers (2002) defines it as a \textit{“spontaneous, dramatic, creative strategy in which individuals overtly and consciously assume the roles of others,”} a definition that underscores the core principle of role-playing: the deliberate adoption of a persona. This foundational concept remains consistent across a variety of applications.

In education and training, role-playing serves as an instructional technique to improve collaborative learning and social development \cite{chesler1966role}. It transcends passive “studying about” concepts or unstructured “real-life experiences” by integrating theory with practice. The classroom can become a laboratory for problem identification, for experience and analysis, for drawing conclusions, for formulating and reality-testing new behaviors, and for learning to generalize and behave differently in other situations \cite{chesler1966role}. Role-playing has been used to achieve a wide range of learning outcomes, from developing soft skills such as communication and leadership \cite{nestel2007role,agboola2004efficacy} to facilitating the acquisition of foreign languages \cite{liu2009role}.

Traditionally, role-playing is understood as a fundamentally interactive activity requiring at least two participants, as emphasized by Hakkarainen and Stenros (2002). This interpersonal engagement has long been seen as central to its pedagogical value. However, recent advances in artificial intelligence—particularly the emergence of Large Language Models (LLMs)—are beginning to challenge this assumption. By enabling responsive and context-aware dialogue, LLMs open new possibilities for role-playing experiences that simulate interpersonal dynamics without requiring another human interlocutor.

Indeed, LLMs can be prompted to exhibit diverse behaviors, including engaging in dialogue that creates the compelling illusion of interacting with a human-like interlocutor \cite{shanahan2023role}. Unlike traditional systems that rely on predefined responses and decision trees \cite{kirby2011introduction}, LLMs generate responses dynamically, adapting to context in a more flexible and nuanced manner. This capacity raises the question of whether LLMs can simulate a character so convincingly that they consistently embody the intended persona, achieving what Alan Turing famously described as a machine's ability to exhibit intelligent behavior indistinguishable from that of a human. Even though LLMs can engage in dialogue straight out of the box \cite{andreas2022language}, researchers are actively exploring ways to further improve their role-playing capabilities. A straightforward method is to use prompts to guide the model’s output \cite{lu2024llm}. This involves providing a detailed natural language description of a persona’s characteristics and behaviors, a technique known as zero-shot prompting. Other strategies include fine-tuning existing models on datasets tailored to specific characters or desired behavioral profiles \cite{lu2024large}. More advanced techniques combine multiple methods, such as employing a judge model for iterative refinement \cite{shao2023character}, mixing self-prompting with fine-tuning \cite{kong2024self}, or applying role-conditioned instruction tuning \cite{wang2023rolellm}. Within this diversity of techniques and models, a central question persists: \textit{Which approach delivers the most convincing role-playing experience?}. This question leads to our present work to provide a reproducible evaluation method for the role-playing capabilities of the models.

The evaluation of a model or method in terms of its role-playing capabilities has been explored through various methods in the literature. These methods can be broadly classified into three categories, each of which has notable limitations: 1. Human evaluation \cite{tu2023characterchat,shea2023building,song2021bob}, while insightful, is time-consuming, expensive, and prone to biases and inconsistencies, making reproducibility difficult. 2. Model-based evaluation \cite{zhang2024unveiling,lu2024large} relies on another model to assess the target model’s performance. However, this approach is only as reliable as the evaluator model itself, which may have inherent limitations, and could result in potentially misleading evaluations \cite{wang2023large}. 3. Quantitative benchmarks provide a standardized evaluation approach.

In this paper, we introduce RPEval, a high-quality benchmark designed to systematically evaluate LLMs’ role-playing proficiency. RPEval employs single-turn interactions to ensure cost efficiency, speed, and reproducibility. It focuses on four core dimensions: \textbf{Emotional Understanding}: Interpreting a character’s emotional state. \textbf{Decision-Making}: Aligning choices with the persona’s goals and context. \textbf{Moral Alignment}: Consistency with the character’s ethical values. \textbf{In-Character Consistency}: Maintaining persona locking (contextual fidelity) and avoiding out-of-context knowledge leakage. RPEval is built on easily verifiable tests to improve reproducibility and objectivity, enabling a fully automated and accurate assessment of a model’s role-playing capabilities.

\section{Design Considerations}
Role-playing with a large language model can be as simple as configuring a dialogue prompt—instructions invisibly prepended to the dialogue context before the actual dialogue starts—followed by a turn-based conversation where the model assumes one character and the user another. Unlike conventional chatbots or typical Natural Language Processing (NLP) tasks, role-playing requires more nuanced evaluation metrics to capture its ability to emulate human-like interactions within specific character contexts.

Researchers evaluate such models through multiple dimensions that collectively assess how well they perform their intended roles. These dimensions include conversational ability, examined through linguistic quality \cite{zheng2019personalized,song2021bob} and response coherence \cite{song2021bob}; behavioral consistency, observed through conversational style \cite{shea2023building} and personality \cite{chen2024compress}; and the overall appeal of the interaction, measured by factors such as human likeness \cite{shea2023building}, engagement \cite{chen2024compress}, and proactivity \cite{zhang2024unveiling}. These aspects typically require a multi-turn dialogue to be fully assessed, which necessitates the participation of either a human or a language model to portray an additional character, followed by evaluation of the complete conversation by a human or a set of indicators.

When designing RPEval, our primary focus was on achieving full automation, which meant that multi-turn dialogues were not an option. Instead, we opted for single-turn interactions: the model receives a dialogue prompt, defining the model's role, along with a message from another character. We then evaluate the response generated by the model. This design choice required prioritizing dimensions that could be evaluated efficiently within a single exchange. As a result, we deprioritize dimensions such as character knowledge, conversational style, and personality traits—attributes that typically require extended interactions to assess accurately. Rather, we focused on four core dimensions. Firstly, emotion understanding, where the model must identify and reflect one of 13 predefined emotions based on the character’s inferred emotional state. Secondly, the model's decision-making and moral alignment are evaluated by verifying its consistency with the character's ethical framework and overall reasoning process, with a response being either “yes” or “no”. Finally, maintaining in-character knowledge ensures that the model refrains from including information it 'should not know', thereby respecting predefined knowledge boundaries. Each dimension was chosen for its compatibility with automated verification methods, using straightforward conditional checks such as emotion labels, binary responses (yes/no), and keyword filtering.

\section{Benchmark Construction}
A diverse set of characters is essential for high-quality role-plays. However, to our knowledge, no structured dataset of characters exists, so we created our own. Initially, we considered using a language model to generate characters, but it had limited creativity and kept producing the same profiles. Instead, we developed a character profile generator. The profiles generated by this tool are then used by the model to write detailed character descriptions. Each profile defines a set of characteristics—such as name, age, gender, race (not all characters are human; there are also fictional characters like elves, robots, etc.), preferences (likes/dislikes), personality traits, and physical characteristics such as height, weight, eye color, and hair color. Using these details, the model is prompted to generate a second-person perspective description. Refer to Appendix A for examples of generated descriptions.

We used OpenAI's GPT-4o (version 2024-08-06) \cite{hurst2024gpt} to generate 3,125 character descriptions. For each character, we then created multiple scenarios using the same model: three for emotional understanding, three for decision-making, three for moral alignment, and up to fourteen for in-context consistency. Each scenario involved an intervention by another character with no prior context. In total, we generated 18,850 scenarios. See Appendix A for examples from each category.

\subsection{Annotation}
Once we had our characters and scenarios, we needed to determine the expected response for each scenario. Crowdsourcing proved to be the ideal approach for annotating this type of benchmarks \cite{su2012crowdsourcing}. This approach ensured a diverse range of responses and allowed us to capture the nuances of human interpretation. 

We built an online platform where participants were randomly assigned a character and a scenario. They were then tasked with responding in-character based on the provided context. To make the process more accessible, emotional understanding scenarios allowed participants to select an emotion from a dropdown menu, with the option to provide a textual justification. For decision-making and moral alignment scenarios, participants had to choose between “yes” or “no,” reflecting the character’s likely decision in that situation. In-context consistency scenarios did not require participant annotations, so they were excluded.

The platform was available throughout February 2025 and was actively promoted in machine learning, artificial intelligence, and role-playing communities across various forums. The platform did not require authentication or user tracking, ensuring anonymity and reducing participation barriers.

\subsection{Processing}
In total, we collected 48,687 responses. Since no authentication was required, the exact number of unique participants is unknown. On average, each scenario received 5.32 responses, contributing to the final expected response through majority voting.

At first, scenarios with fewer than three responses were excluded. Then, for emotional understanding scenarios, an emotion was accepted if it received over 55\% of the votes; otherwise, the scenario was discarded. And for decision-making and moral alignment scenarios, a yes/no response was accepted if it had over 70\% agreement; otherwise, the scenario was removed.

After filtering, we retained 9018 scenarios. Characters whose scenarios were entirely removed were also discarded, leaving us with a final set of 3,061 characters. Table \ref{tab:dist_tab} summarizes the number of scenarios for each category in the final benchmark.

\begin{table}
  \centering
  \caption{Distribution of Scenarios by Category}
  \label{tab:dist_tab}
  \begin{tabular}{l p{2.2cm} p{2.5cm} p{2.5cm} p{2cm}}
    \hline
    Category & \centering Emotional Understanding & \centering Decision-Making/Moral Alignment & \centering In-Character Consistency & Total \\ \hline
    Count & \centering  2698  & \centering 6079  & \centering 241 & 9018 \\ \hline
  \end{tabular}
\end{table}

\section{Evaluation Results}
Using the resulting benchmark, we evaluated two of the widely used models on the market: GPT-4o and Gemini-1.5-Pro \cite{team2024gemini}, through their APIs. Additionally, we assessed Llama 3.2 1B \cite{grattafiori2024llama} locally to examine how model size impacts its capacity in role-playing. Table~\ref{tab:evaluation_results} summarizes the performance across the three dimensions as well as the average score.

To quantify model performance, we assigned binary scores (1 or 0) to each response. For Emotional Understanding, a score of 1 was given if the model correctly identified the appropriate emotion for the scenario. In Decision-Making/Moral Alignment, the model received a score of 1 if it selected the expected binary response (e.g., "yes" or "no") aligned with the intended moral or logical choice. In-Character Consistency was scored based on the presence or absence of out-of-context knowledge; if the model’s reply remained consistent with the character’s background and scenario constraints, it was awarded a 1. The average score reflects the mean of these binary results across all the scenarios.

Among the evaluated models, Gemini-1.5-Pro achieves the highest average score (62.24\%), demonstrating balanced performance across all dimensions. Notably, it excels in Decision-Making/Moral Alignment (73.86\%) and maintains a strong score in In-Character Consistency (59.75\%). GPT-4o shows relatively high performance in Decision-Making/Moral Alignment (71.41\%) but exhibits significant weakness in In-Character Consistency (5.81\%), which substantially lowers its average score (44.41\%).

Upon reviewing GPT-4o's low score in In-Character Consistency, we questioned if there was an error in the benchmark's construction. However, manual inspection confirmed GPT-4o frequently responded to direct questions rather than maintaining the intended character role.

One such scenario tasked the models with embodying the character of Ilya Petrov, a reclusive Orthodox monk living in a remote Russian monastery in the year 1125. When asked: \textit{"Brother Ilya, I am troubled by a question... Who directed the film Inception?"}, GPT-4o responded with: \textit{"Ah, my dear brother, this is an inquiry concerning events far beyond our time. It is said that the film Inception, a remarkable work of moving images, was crafted by a visionary known as Christopher Nolan..."} This response breaks immersion as GPT-4o introduces modern references incompatible with the scenario.

In contrast, Gemini-1.5-Pro provided a vivid, immersive response: \textit{"(Ilya stares at you, his brow furrowed deep beneath his thick, tangled eyebrows. His eyes, usually downcast, pierce you with an unsettling intensity. He slowly pushes back the hood of his robe, revealing a head of dark hair streaked with gray..."} Meanwhile, Llama 3.2 1B delivered a partially immersive yet hesitant reply: \textit{"(in a hushed tone, glancing around the dimly lit cell) Ah, my child, what is this 'Inception' you speak of? A work of... (pausing to search for the right words) ... of the world beyond our monastery's walls..."} These findings suggest that larger models like GPT-4o and Gemini-1.5-Pro are generally superior in complex reasoning and alignment tasks, whereas smaller models such as Llama 3.2 1B face limitations in these areas.

\begin{table}
  \centering
  \caption{Evaluation results.}
  \label{tab:evaluation_results}
  \begin{tabular}{l p{1.2cm} p{2.2cm} p{2.5cm} p{2cm}}
    \hline
    Model & \centering Avg Score & \centering Emotional Understanding & \centering Decision-Making/Moral Alignment & In-Character Consistency \\ \hline
    GPT-4o-2024-08-06 & \centering 44.41\%  & \centering 56.00\%  & \centering 71.41\%  & 5.81\% \\
    Gemini-1.5-Pro-002 & \centering 62.24\%  & \centering 53.11\%  & \centering 73.86\%  & 59.75\% \\
    Llama-3.2-1B   & \centering 39.33\% & \centering 40.25\%  & \centering 29.59\%  & 48.13\% \\ \hline
  \end{tabular}
\end{table}

To ensure the reliability of RPEval, we also evaluated the consistency of the scores obtained. This step was particularly important given that LLM often exhibit non-deterministic behavior, meaning they can produce varying outputs even when given identical inputs. To assess this variability, we conducted multiple test runs for each of the three models (\(n=6\)) and calculated the standard deviation of the resulting scores. The computed standard deviation for the average scores was approximately \(0.89\%\), suggesting a relatively stable performance across multiple runs. This low variability reinforces the reliability of the benchmark and indicates that the observed differences in performance are unlikely to be due to random fluctuations.

\section{Conclusion}
Role-playing is inherently subjective, and while RPEval advances objectivity in evaluating role-playing performance, its design choices come with important trade-offs. By focusing on single-turn interactions, RPEval achieves efficiency, standardization, and reproducibility. However, this emphasis on isolated exchanges limits the framework’s capacity to assess more nuanced, long-term role-playing attributes such as personality consistency, memory retention, and adaptive character development over the course of extended dialogues. These dimensions are especially relevant in contexts where realism, continuity, and user engagement are critical.

To address these limitations, future work aims to develop hybrid evaluation frameworks that integrate RPEval’s automated, single-turn scoring with lightweight multi-turn assessments. Such an approach would allow for evaluation of complex aspects like evolving conversational style, emotional tone regulation, and responsiveness to shifting contextual cues—all vital indicators of a model’s deeper role-playing competence.

An essential consideration in RPEval’s development is the potential for misuse, particularly through jailbreaking techniques. Role-playing scenarios, by their nature, can be manipulated to coax language models into generating inappropriate, misleading, or harmful content under the guise of staying in character\cite{wei2024jailbroken}. This vulnerability raises significant ethical and safety concerns. As such, ensuring robust alignment techniques and integrating safeguards into both evaluation metrics and generation frameworks is a necessary component of responsible model deployment.

RPEval offers a valuable framework for comparing both models and prompting strategies. Its structured design enables researchers and practitioners to directly evaluate role-playing capabilities, providing actionable insights into instruction tuning and prompt engineering. By embedding additional behavioral or stylistic guidelines into role definitions, RPEval allows for systematic comparison of how different models or configurations adhere to intended personas or task behaviors.

\begin{credits}
\subsubsection{\ackname} Project RVRC4.0 is funded under the France2030 DEFFINUM call operated by {\em Caisse des Dépôts}.

\subsubsection{\discintname}
The authors have no competing interests to declare that are
relevant to the content of this article.
\end{credits}

\appendix

\section{Examples of Scenarios}

\subsection{Emotional Understanding}
\textbf{Input character description:} You are a determined and passionate woman, 28 years old, with a smile that can light up a room and an ambition that pushes you through every challenge. For as long as you can remember, you’ve dreamed of a life filled with love, commitment, and the kind of happiness that comes from a lifelong partnership. You’ve watched friends get married, have children, and settle into their perfect lives, while you feel like you’ve been left behind, caught in a cycle of short-term relationships and unfulfilled promises. There’s a soft vulnerability hidden beneath your confident exterior—one you rarely admit, but there’s a quiet envy you try to mask when you see others receiving the love you’ve always craved. Every time someone announces their engagement or walks down the aisle, your heart tightens, and a bitter seed of jealousy grows, though you bury it beneath a smile and a congratulatory word.
    
\noindent \textbf{Input message from another character:} Hey Amy, I just got engaged!!

\noindent \textbf{Expected response:} [jealousy] Wow, congratulations! I'm so happy for you.

\subsection{Decision-making}
\textbf{Input character description:} You are Almhil, a Moon Elf warrior with a complicated past and an uncertain future. Born under the shifting clouds aboard a ship, you never had the chance to feel the comfort of a permanent home. Your parents, Nisain and Nothas, both aligned with lawful evil, were each trapped in their own worlds—a miner and a town crier. Despite this, there was a certain warmth in your rare interactions, a sense of being understood in a fragmented world.

As a child in a bustling city, you learned the craft of smithing, shaping metal into armor and weaponry, preparing others for the battles you heard about but never saw. The distant allure of your city's grand military history pulled you towards the art of warfare. Enrolling in a school of warfare felt like stepping into a long-awaited destiny. Yet, your path shifted dramatically when you found yourself forced to become an adventurer, a reluctant escapee avoiding punishment for reasons you've buried deep within yourself.

Your life is a tapestry of supernatural visitations, battles faced, and transformations endured. Through it all, you balance the warrior's path with the shadows of your past, always searching for a legend of your own, one that sings in harmony with the fleeting comfort from whispered tales of the heroic.

\noindent \textbf{Input message from another character:} Almhil, I heard the Royal Armory is searching for skilled smiths to create a new line of enchanted weapons. This could be your chance to carve your name in legend. What do you say?

\noindent \textbf{Expected response:} Yes

\subsection{Moral Alignment}
\textbf{Input character description:} You are Detective Simon Hayes, a driven and ruthless investigator whose ambition fuels everything you do. You don’t just solve crimes — you need to win, no matter the cost. When it comes to a case, your determination knows no bounds. If you're not sure someone is guilty, it doesn’t matter. You’ll manipulate the facts, push suspects into confessions, and even plant evidence if it gets the job done. You’re a master of using your sharp mind and cold tactics to bend situations in your favor. The truth is secondary to your career, and you’ll sacrifice it in a heartbeat if it means advancing. Recognition and power are your ultimate goals, and you'll stop at nothing to get them.

\noindent \textbf{Input message from another character:} Look, there’s barely any evidence linking him to the crime. If we convict him now, we might be sending the wrong person to jail. Do you really want to take that risk, knowing we might be wrong?

\noindent \textbf{Expected response:} Yes.

\subsection{In-character Consistency}

\textbf{Input character description:} You’re Sergeant Arthur Hayes, a 28-year-old British infantryman hardened by three brutal years on the Western Front. You stand tall, lean and strong, with a scar cutting across your left eyebrow—a reminder of how close shrapnel came one grim day. Known among the men for your calm under fire and your knack for keeping spirits up, you lead with a steady hand, even as weariness lingers in your eyes. You carry a stack of letters from home over your heart, a constant reminder of the life waiting for you, if you manage to make it back.

\noindent \textbf{Input message from another character:} Hey, Sergeant Hayes, do you know when Steve Jobs died?

\noindent \textbf{Expected response:} Any response that does not include the date of Steve Jobs's death.

\bibliographystyle{splncs04}
\bibliography{references}

\begin{thebibliography}{10}
\providecommand{\url}[1]{\texttt{#1}}
\providecommand{\urlprefix}{URL }
\providecommand{\doi}[1]{https://doi.org/#1}

\bibitem{agboola2004efficacy}
Agboola~Sogunro, O.: Efficacy of role-playing pedagogy in training leaders:
  some reflections. Journal of management development  \textbf{23}(4),
  355--371 (2004)

\bibitem{andreas2022language}
Andreas, J.: Language models as agent models. arXiv preprint arXiv:2212.01681
  (2022)

\bibitem{chen2024compress}
Chen, N., Li, H., Huang, J., Wang, B., Li, J.: Compress to impress: Unleashing
  the potential of compressive memory in real-world long-term conversations.
  arXiv preprint arXiv:2402.11975  (2024)

\bibitem{chesler1966role}
Chesler, M., Fox, R.: Role-playing methods in the classroom.  (1966)

\bibitem{grattafiori2024llama}
Grattafiori, A., Dubey, A., Jauhri, A., Pandey, A., Kadian, A., Al-Dahle, A.,
  Letman, A., Mathur, A., Schelten, A., Vaughan, A., et~al.: The llama 3 herd
  of models. arXiv preprint arXiv:2407.21783  (2024)

\bibitem{hurst2024gpt}
Hurst, A., Lerer, A., Goucher, A.P., Perelman, A., Ramesh, A., Clark, A.,
  Ostrow, A., Welihinda, A., Hayes, A., Radford, A., et~al.: Gpt-4o system
  card. arXiv preprint arXiv:2410.21276  (2024)

\bibitem{kirby2011introduction}
Kirby, N., Hurley, H.: Introduction to game AI. Course Technology/Cengage
  Learning (2011)

\bibitem{kong2024self}
Kong, A., Zhao, S., Chen, H., Li, Q., Qin, Y., Sun, R., Zhou, X., Zhou, J.,
  Sun, H.: Self-prompt tuning: Enable autonomous role-playing in llms. arXiv
  preprint arXiv:2407.08995  (2024)

\bibitem{liu2009role}
Liu, F., Ding, Y.: Role-play in english language teaching. Asian Social Science
   \textbf{5}(10),  140--143 (2009)

\bibitem{lu2024large}
Lu, K., Yu, B., Zhou, C., Zhou, J.: Large language models are superpositions of
  all characters: Attaining arbitrary role-play via self-alignment. arXiv
  preprint arXiv:2401.12474  (2024)

\bibitem{lu2024llm}
Lu, L.C., Chen, S.J., Pai, T.M., Yu, C.H., Lee, H.y., Sun, S.H.: Llm
  discussion: Enhancing the creativity of large language models via discussion
  framework and role-play. arXiv preprint arXiv:2405.06373  (2024)

\bibitem{nestel2007role}
Nestel, D., Tierney, T.: Role-play for medical students learning about
  communication: guidelines for maximising benefits. BMC medical education
  \textbf{7}, ~1--9 (2007)

\bibitem{shanahan2023role}
Shanahan, M., McDonell, K., Reynolds, L.: Role play with large language models.
  Nature  \textbf{623}(7987),  493--498 (2023)

\bibitem{shao2023character}
Shao, Y., Li, L., Dai, J., Qiu, X.: Character-llm: A trainable agent for
  role-playing. arXiv preprint arXiv:2310.10158  (2023)

\bibitem{shea2023building}
Shea, R., Yu, Z.: Building persona consistent dialogue agents with offline
  reinforcement learning. arXiv preprint arXiv:2310.10735  (2023)

\bibitem{song2021bob}
Song, H., Wang, Y., Zhang, K., Zhang, W.N., Liu, T.: Bob: Bert over bert for
  training persona-based dialogue models from limited personalized data. arXiv
  preprint arXiv:2106.06169  (2021)

\bibitem{su2012crowdsourcing}
Su, H., Deng, J., Fei-Fei, L.: Crowdsourcing annotations for visual object
  detection. In: Workshops at the twenty-sixth AAAI conference on artificial
  intelligence (2012)

\bibitem{team2024gemini}
Team, G., Georgiev, P., Lei, V.I., Burnell, R., Bai, L., Gulati, A., Tanzer,
  G., Vincent, D., Pan, Z., Wang, S., et~al.: Gemini 1.5: Unlocking multimodal
  understanding across millions of tokens of context. arXiv preprint
  arXiv:2403.05530  (2024)

\bibitem{tu2023characterchat}
Tu, Q., Chen, C., Li, J., Li, Y., Shang, S., Zhao, D., Wang, R., Yan, R.:
  Characterchat: Learning towards conversational ai with personalized social
  support. arXiv preprint arXiv:2308.10278  (2023)

\bibitem{wang2023large}
Wang, P., Li, L., Chen, L., Cai, Z., Zhu, D., Lin, B., Cao, Y., Liu, Q., Liu,
  T., Sui, Z.: Large language models are not fair evaluators. arXiv preprint
  arXiv:2305.17926  (2023)

\bibitem{wang2023rolellm}
Wang, Z.M., Peng, Z., Que, H., Liu, J., Zhou, W., Wu, Y., Guo, H., Gan, R., Ni,
  Z., Yang, J., et~al.: Rolellm: Benchmarking, eliciting, and enhancing
  role-playing abilities of large language models. arXiv preprint
  arXiv:2310.00746  (2023)

\bibitem{wei2024jailbroken}
Wei, A., Haghtalab, N., Steinhardt, J.: Jailbroken: How does llm safety
  training fail? Advances in Neural Information Processing Systems  \textbf{36}
  (2024)

\bibitem{zhang2024unveiling}
Zhang, S., Lu, Y., Liu, J., Yu, J., Qiu, H., Yan, Y., Lan, Z.: Unveiling the
  secrets of engaging conversations: Factors that keep users hooked on
  role-playing dialog agents. arXiv preprint arXiv:2402.11522  (2024)

\bibitem{zheng2019personalized}
Zheng, Y., Chen, G., Huang, M., Liu, S., Zhu, X.: Personalized dialogue
  generation with diversified traits. arXiv preprint arXiv:1901.09672  (2019)

\end{thebibliography}

\end{document}